\documentclass[10pt,twocolumn,letterpaper]{article}

\usepackage[pagenumbers]{cvpr} 

\usepackage{graphicx}
\usepackage{amsmath}
\usepackage{amssymb}
\usepackage{booktabs}
\usepackage{multirow}
\usepackage{tabularx}
\usepackage[mathscr]{eucal}
\usepackage{colortbl}
\definecolor{mygray}{gray}{.9}
\usepackage{lipsum}
\usepackage[hang,flushmargin]{footmisc}

\usepackage[pagebackref,breaklinks,colorlinks]{hyperref}

\usepackage[capitalize]{cleveref}
\crefname{section}{Sec.}{Secs.}
\Crefname{section}{Section}{Sections}
\Crefname{table}{Table}{Tables}
\crefname{table}{Tab.}{Tabs.}

\def\cvprPaperID{9168} 
\def\confName{CVPR}
\def\confYear{2023}

\begin{document}

\title{Boundary-aware Backward-Compatible Representation via Adversarial Learning in Image Retrieval}

\author{\normalsize{Tan Pan$^*$, Furong Xu$^*$$^\dag$, Xudong Yang$^\dag$, Sifeng He, Chen Jiang, Qingpei Guo, 
Feng Qian,} \\
\normalsize{Xiaobo Zhang, Yuan Cheng, Lei Yang, Wei Chu} \\
Ant Group\\
{\tt\small {\{pantan.pt$^*$, booyoungxu.xfr$^*$$^\dag$, jiegang.yxd$^\dag$\}}@antgroup.com}
}
\maketitle
\newcommand\blfootnote[1]{%
\begingroup
\renewcommand\thefootnote{}\footnote{#1}%
\addtocounter{footnote}{-1}%
\endgroup
}
\blfootnote{
    *These authors contributed equally to this research.\\
    {\dag}Corresponding authors.
}

\begin{abstract}
   Image retrieval plays an important role in the Internet world. Usually, the core parts of mainstream visual retrieval systems include an online service of the embedding model and a large-scale vector database. For traditional model upgrades, the old model will not be replaced by the new one until the embeddings of all the images in the database are re-computed by the new model, which takes days or weeks for a large amount of data. Recently, backward-compatible training (BCT) enables the new model to be immediately deployed online by making the new embeddings directly comparable to the old ones. For BCT, improving the compatibility of two models with less negative impact on retrieval performance is the key challenge. In this paper, we introduce AdvBCT, an Adversarial Backward-Compatible Training method with an elastic boundary constraint that takes both compatibility and discrimination into consideration. We first employ adversarial learning to minimize the distribution disparity between embeddings of the new model and the old model. Meanwhile, we add an elastic boundary constraint during training to improve compatibility and discrimination efficiently. Extensive experiments on GLDv2, Revisited Oxford (ROxford), and Revisited Paris (RParis) demonstrate that our method outperforms other BCT methods on both compatibility and discrimination. The implementation of AdvBCT will be publicly available at \href{https://github.com/Ashespt/AdvBCT}{https://github.com/Ashespt/AdvBCT}.
\end{abstract}

\section{Introduction}
\label{sec:intro}
\begin{figure*}
  \centering
  \begin{subfigure}{0.47\linewidth}
    \centering
    \includegraphics[width=1.05\textwidth]{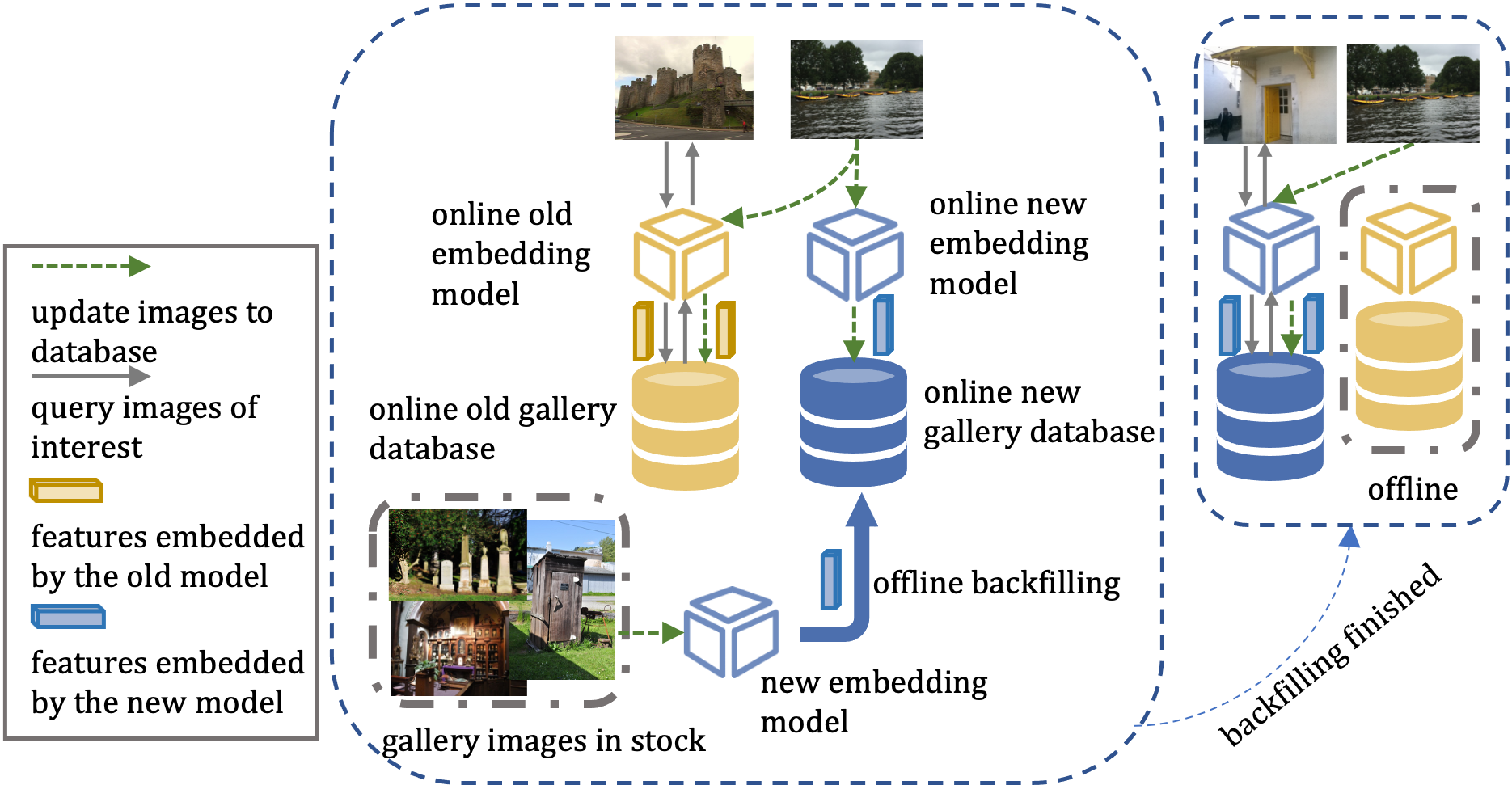}
    \caption{The process of the traditional upgrade.}
    \label{fig:process-a}
  \end{subfigure}
  \hspace{-5mm}
  \begin{subfigure}{0.47\linewidth}
    \centering
    \includegraphics[width=0.8\textwidth]{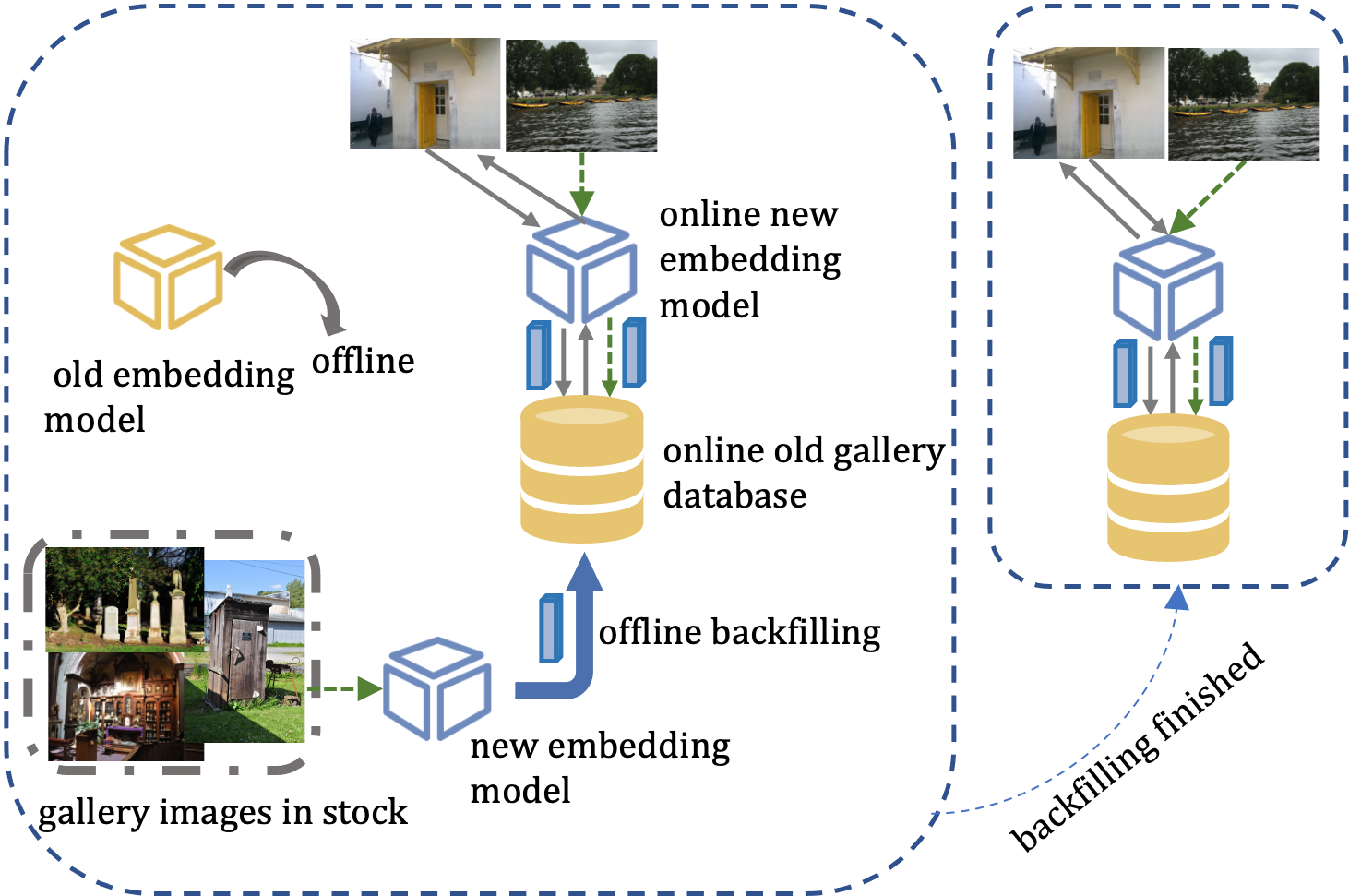}
    \caption{The process of the compatible upgrade.}
    \label{fig:process-b}
  \end{subfigure}
  \caption{The processes of updating online systems on different models. \textbf{Best viewed in color.}}
  \label{fig:process}
\end{figure*}

Image retrieval brings great convenience to our daily life in various areas such as e-commerce search\cite{rowley2000product,li2011towards}, face recognition \cite{zhao2003face,schroff2015facenet}, and landmark localization \cite{xu20213rd,mei20203rd}. With the rapid development of deep learning, visual retrieval systems develop towards larger models and richer databases to provide people with better services. Most modern visual retrieval systems include two core parts: (i) an online service of the embedding model which maps an input image to a high-dimensional vector, i.e., the embedding, and (ii) a large-scale vector database which stores the embeddings of the gallery set and is responsible for similarity search when a query image arrives. During the lifetime of the system, new models with better performance are trained and then deployed online to replace the old ones. Unfortunately, the embeddings of the query images extracted by the new model are not compatible with the old ones in the database in most cases. As a result, the vector database must be rebuilt by extracting embeddings for the whole gallery set with the new model. This process is called backfilling \cite{shen2020towards}. 
In general, practical industrial applications containing a database with millions to billions of images take days or even weeks for backfilling. During that time, the old model and database must be kept online to handle queries. This so-called cold-refresh \cite{zhang2021hot} model upgrade process is shown in \cref{fig:process-a}.

Recently, to save resources and simplify the complex backfilling process, backward-compatible learning was proposed \cite{shen2020towards}. Backward-compatible learning aims to ensure the compatibility of embedding representations between models. As shown in \cref{fig:process-b}, the new model can directly replace the old one and the embeddings of images in stock are updated on-the-fly. With the hot-refresh model-update strategy, the system only needs to maintain one service and one database at a time, effectively reducing the required resource during backfilling.
\begin{figure}
  \centering
    \includegraphics[width=0.7\linewidth]{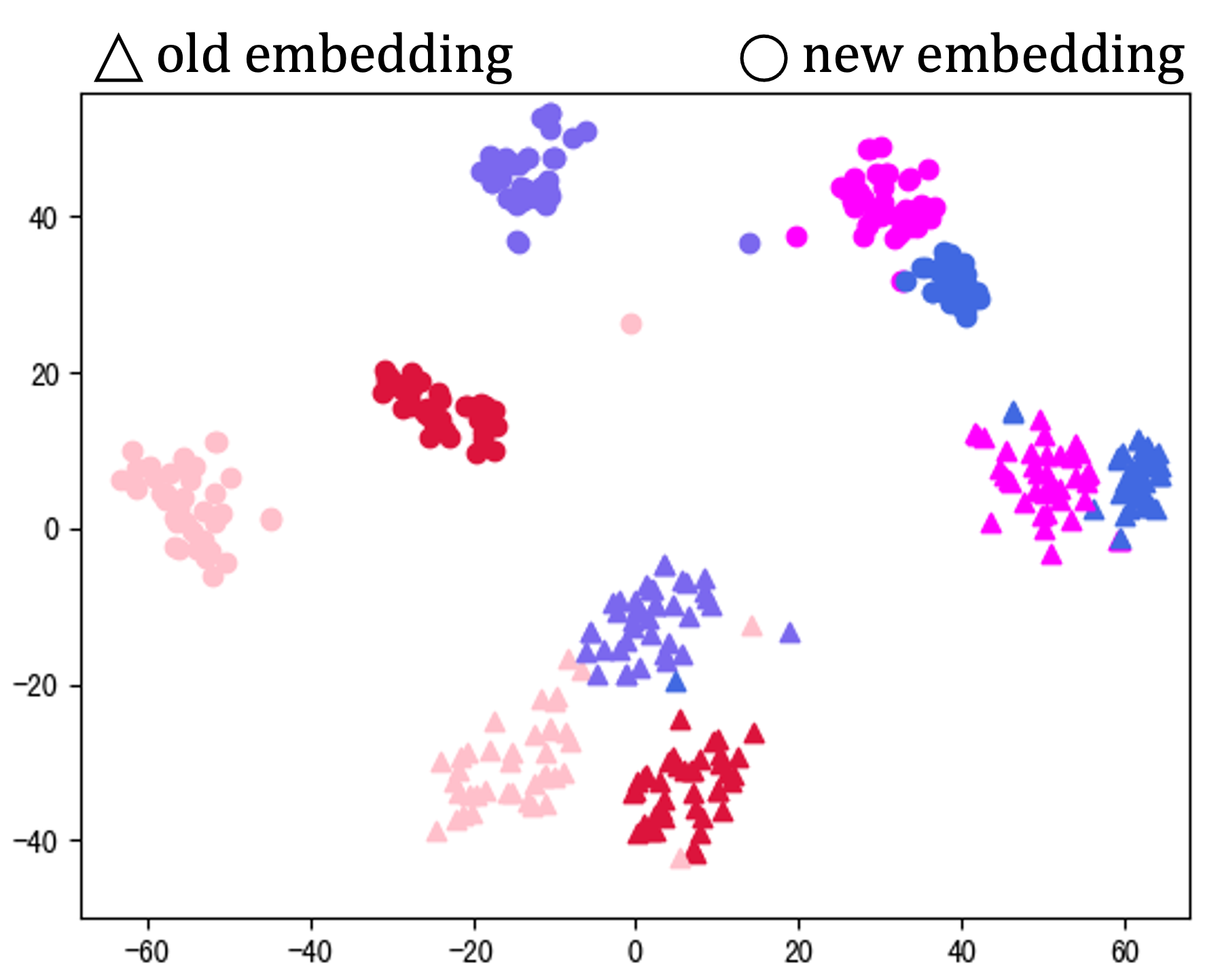}
  \caption{Distributions visualization of the old embeddings and new embeddings on RParis without compatible training by t-SNE \cite{van2008visualizing}. Triangles represent old embeddings and circles represent new embeddings. The old embeddings are extracted by the model trained on 30\% data of GLDv2 while the new embeddings extracted by the model trained on 100\% data of GLDv2. The embeddings in the same color belong to same class.}
  \label{fig:contribution}
\end{figure}

Meanwhile, these hot-refresh model upgrades also pose new challenges for visual search systems. The retrieval performance during backfilling reflects the compatibility between models, which should not degrade compared to the old system. However, making the new model perfectly back-compatible with the old one leaves no room for improvement on the retrieval task. The ultimate goal of the compatible learning is to enable the compatibility between the new model and the old model while keeping the performance gain of the back-compatible trained new model as close as possible to that of the independently trained new model. When evaluating BCT methods, both the compatibility between models and the discrimination of the new model for the retrieval task must be taken into account.

The incompatibility between models results from the discrepancy of the embedding distributions of the models, which is illustrated in \cref{fig:contribution}. Most of the previous works in BCT \cite{zhang2022towards,meng2021learning,zhang2021hot} narrow the distribution gap by adding some regularization losses involving the old and the new embeddings. The main idea of these methods is to pull the new and old embeddings of the same class closer and to push the new and old embeddings of different classes apart from each other in a metric learning manner. Another way to minimize the distribution discrepancy is adversarial learning, which was successfully applied in domain adaptation \cite{ganin2015unsupervised,hu2018duplex}. We decide to combine the metric learning and the adversarial learning as they measure and minimize the discrepancy between distributions in different ways, and this is complementary for compatible learning in our intuition.
Some works \cite{zhang2021hot,wu2022neighborhood} design the loss for the compatibility in a point-to-point manner, which is sensitive to outliers in the training data. Other works \cite{meng2021learning,shen2020towards} propose point-to-set losses to address the issues by loosely constraining the new embeddings inside the class boundary estimated by the old model. However, these estimated boundaries remain constant when training the new model, which may not be flexible for the new model to learn more discriminative embeddings. We design an elastic boundary loss in which the boundary can be dynamically adjusted during training.



In addition, existing methods are evaluated in different settings and on different datasets, which makes it difficult to fairly compare them. In this paper, we adopt a unified training and evaluation protocol to evaluate the existing backward-compatible methods and our adversarial backward-compatible training method (\textit{AdvBCT}).

In summary, the main contributions of our work are listed as follows:
\begin{itemize}
	\item We first propose an adversarial backward-compatible learning method to close the distribution gap between different models and employ an elastic boundary loss to improve compatibility and discrimination. 
	\item We unify the training and evaluation protocol to assess the performance of 5 BCT works. Meanwhile, we propose a new metric named $\mathcal{P}_{\beta-score}$ to evaluate compatibility and discrimination in a comprehensive metric.
	\item By comprehensive experiments, we show that the proposed \textit{AdvBCT} outperforms the related state-of-the-arts on several image retrieval datasets, {\it e.g.} GLDv2, ROxford and RParis in most BCT settings.
\end{itemize}

\begin{figure*}
  \centering
    \includegraphics[width=0.8\textwidth]{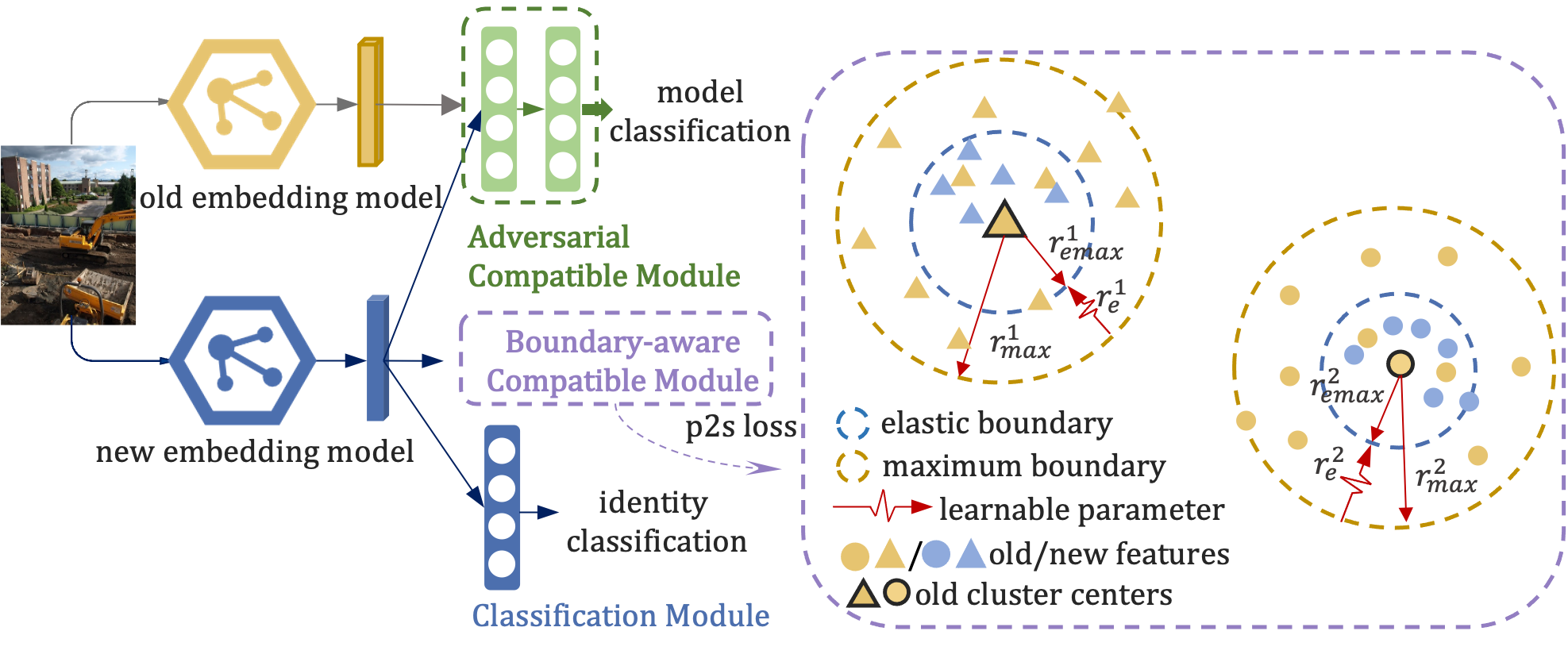}
  \caption{An overview of our \textit{AdvBCT} framework. The adversarial and boundary-aware compatible modules minimize the discrepancy between distributions of the old and new embeddings, while the classification module improves the retrieval performance. In the compatible losses part, the blue and yellow circles refer to the boundary between the new embeddings and old class centers respectively. $r_{max}$ minus the learnable $r_e$ equals to $r_{emax}$. $r_{max}$ is the maximum distance between old embeddings and the old class center. The solid circles and triangles represent embeddings from two different classes.}
  \label{fig:method}
\end{figure*}

\section{Related Works}
\label{sec:rewo}
\noindent\textbf{Backward-Compatible Training.} Aiming to deploy a new model without the operation of ``backfilling", Shen et al. \cite{shen2020towards} proposed backward-compatible learning to get new embeddings compatible with old ones. Following this topic, compatible learning methods have developed into three branches: backward-compatible learning, forward-compatible learning \cite{ramanujan2022forward,zhou2022forward}, and cross-model compatible learning (CMC) \cite{meng2021learning,duggal2021compatibility,srivastava2020empirical,wu2022neighborhood}. Forward-compatible learning aims to enhance the capacity of the current model to leave embedding space for the next upgrade, which is suitable for the close-set scenario. CMC supposes that there are two existing models and adopts projection heads to make models compatible. However, the upgrade of CMC models still needs two embedding models and two databases which is shown in supplemental materials. In this paper, we mainly focus on the backward-compatible training (BCT) situation, which is suitable for open-set scenarios and can simplify the process of backfilling. 

On the BCT track, Wang \etal \cite{meng2021learning} aligned class centers between models by a transformation that can be applied to both cross-model compatibility training and compatible training. Zhang \etal \cite{zhang2022towards} defined 4 types of model upgrading and proposed a novel structural prototype refinement algorithm to adapt to different protocols. Wu \etal \cite{wu2022neighborhood} constrained the relationship between new embeddings and old embeddings inspired by contrastive learning. Those methods proved to be effective under different datasets and protocols and focused on constraints of prototypes and class centers between new and old models.

\noindent\textbf{Image Retrieval.}
Image retrieval is a classic task in representation learning. Given a query image, the system will return the top similar images by searching a large-scale pre-encoded database. Image retrieval is related to many research topics such as landmark retrieval, face recognition, person re-identification, and so on. Researchers pay more attention to the improvement of retrieval performance \cite{xu2020metric,ge2020self,xu2021discrimination,mei20203rd,xu2021prdp} and there exist several comprehensive benchmarks \cite{weyand2020google,maze2018iarpa}. However, those works rarely discuss the model upgrade problem for visual retrieval systems which is a critical concern in real industrial applications.

\noindent\textbf{Adversarial Learning.}
Adversarial learning is widely used in various fields such as generative adversarial networks \cite{goodfellow2020generative,karras2019style,jiang2021transgan}, person re-identity \cite{zheng2019joint}, and domain adaptation \cite{ganin2015unsupervised,hu2018duplex}. Generative adversarial networks utilize a discriminator to enhance the ability of the generator to synthesize samples. Ganin and Lempitsky's work \cite{ganin2015unsupervised} learned embeddings invariant to the domain shift by a label predictor which combined adversarial learning with domain adaptation. After that, various adversarial domain adaptation methods were proposed \cite{lin2019real,long2018conditional}. The core of adversarial learning is minimizing the discrepancy between distributions of target and source by defending against attacks.

\section{Methodology}
\label{sec:method}
\subsection{Problem Formulation}
\label{sec:probfor}
 Given a training dataset $\mathcal{T}=\{X,Y\}$ of $C$ classes, $X=\{x_1,x_2...,x_m\}$ is the image sample set, and $Y=\{y_1,y_2,...,y_m\}$ is the one-hot label set. We can train an embedding model $\phi$ by using $X$ to map image samples into embeddings $Z=\{\phi(x_1),\phi(x_2),...,\phi(x_m)\}$.

We use $\phi_{o}$ trained on $\mathcal{T}_{old}$ and $\phi_{n}$ trained on $\mathcal{T}_{new}$ to represent the old embedding model and new embedding model respectively. Furthermore, $\phi_{*}$ represents the model obtained with the same settings as the new embedding model but without compatibility constraints. In the inference stage, a query dataset $\mathcal{Q}$ and a gallery dataset $\mathcal{G}$ are adopted to evaluate the retrieval performance of embeddings. To measure the compatibility performance during backfilling, the embeddings of $\mathcal{G}$ are extracted by $\phi_{o}$, and the ones of $\mathcal{Q}$ are obtained by $\phi_{n}$.  In this work, $\left \langle \phi(x_i),\phi(x_j)\right \rangle$ represents the distance between two embeddings $\phi(x_i)$ and $\phi(x_j)$ under some distance metrics. The smaller the distance is, the more similar the samples are.

\subsection{Backward-Compatible Training Framework}
The new model obtained by backward-compatible training needs to be compatible with the old model in embedding space, and it is also demanded to be more discriminative than the old model on retrieval performance. To meet these two requirements in model iterations, as shown in figure \cref{fig:method}, we adopt three modules, namely the classification module, the adversarial compatible module and the boundary-aware compatible module. The classification module aims at guiding the new model to learn discriminative embeddings. The adversarial compatible module reduces the discrepancy between the distribution of the old embeddings and the distribution of the new embeddings in an adversarial manner. The boundary-aware compatible module is utilized to maintain a reasonable distance between the new embeddings and the old cluster centers to improve compatibility and discrimination efficiently. These modules are explained in detail in \cref{sec:bcl}.

\subsection{Backward-Compatible Learning}
\label{sec:bcl}

\noindent\textbf{Classification Module.} To ensure the discrimination of new embeddings, we adopt the commonly used method that learn the representation using  close-set classification: a classifier is connected after the embedding layer of the model and cross-entropy loss is employed to minimize the classification error of the new model. In this way, the model can extract discriminative features for similarity retrieval. The cross-entropy loss is defined as:
	\begin{equation}
		\mathcal{L}_{cls} =-\frac{1}{N}\sum_{i=1}^N y_i\log p_i
		\label{equ:cls}
	\end{equation}
where $N$ is the number of instances, $y_i$ is the label of the image $x_i$, and $p_i$ is the probability that $x_i$ belongs to the class $y_i$.


\noindent\textbf{Adversarial Compatible Module.} As mentioned in \cref{sec:intro}, reducing the dissimilarity of the embedding distribution between the new model and the old model is the key for model compatibility. The adversarial learning is applied to domain adaptation \cite{xu2018style,you2019universal,wang2018deep} to guide the embedding model to generate domain-insensitive embeddings. Inspired by this, we adopt adversarial learning to make the embedding distribution between models as similar as possible. Specifically, we introduce a discriminator $\phi_d$ after the embedding layer of the new model as shown in \cref{fig:method}. The discriminator determines whether the embedding comes from the new model or from the old model. Hence, the discrepancy of the embedding distributions is estimated by the classification loss of the discriminator, which is defined as: 
\begin{equation}
	\mathcal{L}_{adv}=E(\theta_n, \theta_d)= -\frac{1}{N}\sum_{i=1}^N \ell_i\log q_i
	\label{equ:adv}
\end{equation}
where $\ell_i$ is a binary label indicating by which model the embedding is generated, and $q_i$ is the probability output by the discriminator. $\theta_n$ and $\theta_d$ are the parameters of $\phi_n$ and $\phi_d$ respectively.

At training time, the embedding model and the discriminator are optimized together in an adversarial way: the model discriminator tries to minimize $\mathcal{L}_{adv}$ while the embedding model tries to maximize it. When the training process converges, the end-to-end network comprised of the embedding model $\phi_n$ and the discriminator $\phi_d$ will look for a saddle point of $\hat{\theta_n}$ and $\hat{\theta_d}$ satisfying:
 	
 \begin{equation}
 \hat{\theta_d}= \mathop{\arg\min}\limits_{\theta_d}	E(\hat{\theta_n}, \theta_d)
 \label{equ:adv_1}
 \end{equation}

\begin{equation}
 	\hat{\theta_n}= \mathop{\arg\max}\limits_{\theta_n}	E(\theta_n, \hat{\theta_d})
 	\label{equ:adv_2}
 \end{equation}
The parameters $\theta_d$ are optimized in the direction of minimizing the classification loss of retrieval as \cref{equ:adv_1} shows. The parameters $\theta_n$ of the new embedding model work on maximizing model classification loss as \cref{equ:adv_2} shows which is equivalent to minimizing the distribution gap.

In order to optimize the end-to-end network composed by the embedding model and the discriminator with opposite objectives using conventional optimization solver such as SGD, a gradient reversal layer (GRL) \cite{ganin2015unsupervised} is inserted between the new embedding model and discriminator. During the forward propagation process, GRL acts as an identity transform. And in the back propagation, GRL takes the gradient from the subsequent level, multiplies it by -$\beta$ and passes it to the preceding layer.



\noindent\textbf{Boundary-aware Compatible Module.} When the new and old embeddings are compatible, the positive pairs and negative pairs should satisfy the following conditions on distance constraints. The following formulas are based on $\forall \{i,j,k\}, y_i=y_j\neq y_k$.
 \begin{equation}
    \left \langle \phi_{n}(x_i),\phi_{o}(x_j) \right \rangle < \left \langle \phi_{n}(x_i),\phi_{o}(x_k) \right \rangle
    \label{equ:con-1}
 \end{equation}
 \begin{equation}
    \left \langle \phi_{n}(x_i),\phi_{o}(x_j) \right \rangle < \left \langle \phi_{n}(x_i),\phi_{n}(x_k) \right \rangle
    \label{equ:con-2}
 \end{equation}

Some methods followed the formulas and designed constraints of new embeddings and old embeddings, which we call point-to-point (\textit{p2p}) constraints. However, the \textit{p2p} constraint is too strict because it is applied to all pairs of samples, which means outliers or corner cases will exert negative effects on training. To address these issues, we transfer this \textit{p2p} constraint into a point-to-set (\textit{p2s}) constraint as follows.

Here, we use Euclidean Metric as the measure of distance $\left \langle \phi_{n}(x_i),\phi_{o}(x_j) \right \rangle = \Vert \phi_{n}(x_i)-\phi_{o}(x_j) \Vert_2$. According to the triangle inequality, we can draw some conclusions on relationships of $\phi_{n}(x_i),\phi_{o}(x_j)$ and $E_o(X^c)$:
\begin{equation}
    B_{lower}=\Vert \phi_{n}(x_i)-E_o(X^c) \Vert_2-\Vert \phi_{o}(x_j)-E_o(X^c) \Vert_2
    \label{equ:dl}
\end{equation}
\begin{equation}
    B_{upper}=\Vert \phi_{n}(x_i)-E_o(X^c) \Vert_2+\Vert \phi_{o}(x_j)-E_o(X^c) \Vert_2
    \label{equ:du}
\end{equation}
\begin{equation}
    B_{lower} \leq \Vert \phi_{n}(x_i)-\phi_{o}(x_j) \Vert_2 \leq B_{upper}
    \label{equ:dldu}
\end{equation}
where $E_o(X^c)$ is the expectation of $\phi_{o}(X^c)$ and $X^c=\{x_i\}_{i=1}^{n}$ is the set of instances of class $\textit{c}$, where $\forall\{x_i,x_j\} \in X^c$. Details are given in the supplemental material Sec. 1.

Because the $\Vert \phi_{o}(x_j)-E_o(X^c) \Vert_2$ is a constant, and the range of $\Vert \phi_{n}(x_i)-\phi_{o}(x_j) \Vert_2$ is determined by $\Vert \phi_{n}(x_i)-E_o(X^c) \Vert_2$. Thus, by constraining the distance between $\phi_{n}(x_i)$ and $E_o(X^c)$, we can constrain the distance between $\phi_{n}(x_i)$ and $\phi_{o}(x_j)$. In this paper, we estimate $E_o(X^c)$ as the cluster center of the training samples for class $c$.

Based on the existing old model, we can easily calculate expectations $\{E_o(X^i)\}_i^C$ namely cluster centers $\mathcal{O} = \{o^1,o^2,...,o^C\}$ of the training data. The maximum distance set of all classes is defined as $R_{max} = \{r_{max}^1,r_{max}^2,...,r_{max}^C\}$ in which $r_{max}^i$ represents the maximum distance between embeddings and the cluster center of class $i$. $r_{max}^i$ is determined by the old embedding space of class $i$. A \textit{p2s} constraint can be $\Vert \phi_{n}(x_i)-E_o(X^i) \Vert_2<r_{max}^i$ in which we call $r_{max}^i$ as the max boundary. However, corner cases and outliers will make the distribution looser which leads to a larger $r_{max}^i$. Although a larger $r_{max}^i$ gives more space to improve discrimination, it will be harder to improve compatibility. We hope to constrain new embeddings and old centers by suitable boundaries. Thus, we use an elastic boundary $r_{e}$ to dynamically adjust the max boundary between a threshold $t$ and $r_{max}$.
$r_{e}$ is defined as:
\begin{equation}
\begin{split}
    r_e^k = w^k\lvert r_{max}^k-t \rvert\\
    \label{equ:re}
\end{split}
\end{equation}
where $w^k$ is a learnable parameter between 0 and 1.

We employ the elastic boundary $r_{e}$ to adjust the final max boundary $r_{emax}$. $r_{emax}$ is defined as \cref{equ:rmax} in which $k$ represents class $k$. The equation will range $r_{emax}$ in $[t,r_{max}]$ or $[r_{max},t]$.
\begin{equation}
\begin{split}
    r_{emax}^k=\left\{
        \begin{array}{lr}
           r_{max}^k-r_e^k   & t<r_{max}\\
           t-r_e^k   &t>r_{max}
        \end{array}
    \right.
    \label{equ:rmax}
\end{split}
\end{equation}

Combined with \cref{equ:rmax} and \cref{equ:re}, the final $r_{emax}$ can be defined as follows:
\begin{equation}
\begin{split}
    r_{emax}^k=\left\{
        \begin{array}{lr}
           (1-w^k)r_{max}^k + w^kt  & t<r_{max}\\
           w^kr_{max}^k+(1-w^k)t   &t>r_{max}
        \end{array}
    \right.
    \label{equ:finalrmax}
\end{split}
\end{equation}

Based on what mentioned above, we design an \textit{elastic p2s loss} to dynamically minimize $\Vert \phi_{n}(x_i)-E_o(X^c) \Vert_2$. As shown in \cref{fig:method}, the \textit{p2s loss} constrains distances between new embeddings and the old cluster center in a dynamic way. The \textit{elastic p2s loss} can be defined as the following:
\begin{equation}
    \mathcal{D}_k=\sum_{i=1}^{m}max(\left \langle \phi_{n}(x_i^k), o^k \right \rangle-r_{emax}^k,0)
    \label{equ:dx2o}
\end{equation}
\begin{equation}
    \mathcal{L}_{p2s}=\sum_{k=1}^{C}\mathcal{D}_k
    \label{equ:p2s}
  \end{equation}

\begin{figure}
  \centering
    \includegraphics[width=0.4\textwidth]{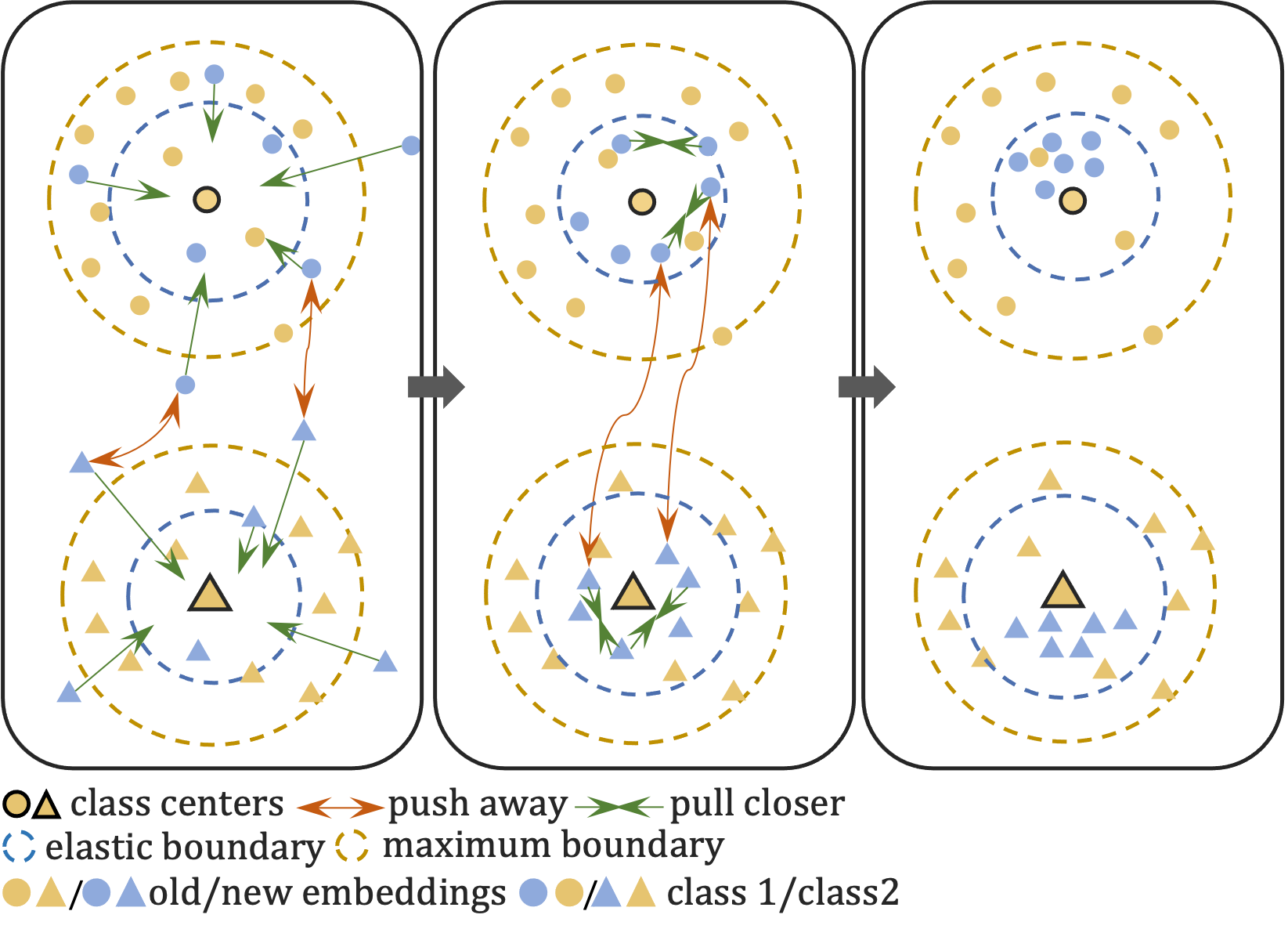}
  \caption{The distribution state evolution of embeddings during training constrained by 
 our $\mathcal{L}_{AdvBCT}$.}
  \label{fig:distanceloss}
\end{figure}
In total, our final loss on embedding model can be summed as the following:
\begin{equation}
	\mathcal{L}_{AdvBCT}=\mathcal{L}_{cls}+\lambda\mathcal{L}_{p2s}+\gamma \mathcal{L}_{adv}
	\label{equ:totalloss}
\end{equation}
where $\gamma$ and $\lambda$ are factors. From the experiments, $\mathcal{L}_{adv}$ affects on the consistency of distributions efficiently in the early training while influences discrimination in the late. So, we progressively reduce $\gamma$ during training. The state of distributions can be abstracted in \cref{fig:distanceloss}.
\section{Benchmarks and Metrics}
\label{sec:benchmark}
\noindent\textbf{Benchmarks.} Training data and backbones can be considered as major factors affecting the performance of the new model when the old model is going to be upgraded. Thus, we discuss three settings as followings. (1) \textbf{Extended-data}: Extended-data supposes that the classes of the new training data set $\mathcal{T}_{new}$ remain unchanged but the data number of each class increases. In our setting, the old training data set $\mathcal{T}_{old}$ composed of $30\%$ images is randomly sampled from the whole data set. (2) \textbf{Extended-class}: $\mathcal{T}_{old}$ is composed of $30\%$ classes of the whole data set. (3) \textbf{Enlarged-backbone}: The old model is trained on ResNet18 (R18) while the new model is trained on ResNet50 (R50). For Extended-data and Extended-class, the new models are trained on R18. Generally, when the backbone is enlarged, the volume of data should be enlarged too. Therefore, in Enlarged-backbone, the data will increase from 30\% to 100\% same as Extended-data or Extended-class.

In this article, except for the settings mentioned above, we won't take situations into account that other training schemes are changed.

\noindent\textbf{Metrics.} Referring to previous works, the performance metrics of compatible learning methods can be divided into two parts, the performance of compatibility during upgrading and the improvement of retrieval after updating which we call discrimination. They can be presented as followings.
\begin{equation}
  \mathcal{P}_{comp}=sigmoid(\frac{M(\phi_{n},\phi_{o};\mathcal{Q},\mathcal{G})-M(\phi_{o},\phi_{o};\mathcal{Q},\mathcal{G})}{M(\phi_*,\phi_*;\mathcal{Q},\mathcal{G})-M(\phi_{o},\phi_{o};\mathcal{Q},\mathcal{G})})
  \label{eq:compatible}
\end{equation}
\begin{equation}
  \mathcal{P}_{up}=sigmoid(\frac{M(\phi_{n},\phi_{n};\mathcal{Q},\mathcal{G})-M(\phi_*,\phi_*;\mathcal{Q},\mathcal{G})}{M(\phi_*,\phi_*;\mathcal{Q},\mathcal{G})})
  \label{eq:upgrading}
\end{equation}
Where $M(\cdot,\cdot)$ represents the evaluation metric for image retrieval, \eg mean Average Precision (mAP). $M(\phi_{n},\phi_{o};\mathcal{Q},\mathcal{G})$ represents the mAP with the setting that embeddings of $\mathcal{Q}$ and embeddings of $\mathcal{G}$ are extracted by $\phi_{n}$ and $\phi_{o}$ respectively.

$\mathcal{P}_{comp}$ is the indicator of compatibility and $\mathcal{P}_{up}$ measures the performance of the new embedding model $\phi_n$ compared to the embedding model  $\phi_*$ trained without compatible methods. We use $sigmoid$ function to normalize values because $M(\phi_{n},\phi_{o};\mathcal{Q},\mathcal{G})$ is less than $M(\phi_{o},\phi_{o};\mathcal{Q},\mathcal{G})$ and $M(\phi_{n},\phi_{n};\mathcal{Q},\mathcal{G})$ is less than $M(\phi_*,\phi_*;\mathcal{Q},\mathcal{G})$ in some cases. For those metrics, a higher value is better. 

Except for the metrics mentioned above, inspired by $F-score$ \cite{matthews1975comparison}, we propose a new metric $P-score$ to measure both the performance of compatibility and discrimination.
\begin{equation}
  \mathcal{P}_{\beta-score}=\frac{(1+\beta^2)\mathcal{P}_{comp}*\mathcal{P}_{up}}{\beta^2\mathcal{P}_{comp}+\mathcal{P}_{up}}
  \label{eq:betascore}
\end{equation}
$P-score$ can take $\mathcal{P}_{comp}$ and $\mathcal{P}_{up}$ into consideration together where $\beta$ is the impact factor of $\mathcal{P}_{comp}$.  Like widely used setting $\beta=1$ in $F-score$, we also set $\beta=1$ in $P-score$ as the formula \cref{eq:pscore} shows. In this case, $P-score$ is the harmonic mean of $\mathcal{P}_{comp}$ and $\mathcal{P}_{up}$.

\begin{equation}
  \mathcal{P}_{1-score}=\frac{2\mathcal{P}_{comp}*\mathcal{P}_{up}}{\mathcal{P}_{comp}+\mathcal{P}_{up}}
  \label{eq:pscore}
\end{equation}
\begin{table}\small
	\centering
	\begin{tabular}{p{2.5cm}<{\centering}|p{0.8cm}<{\centering}p{0.8cm}<{\centering}p{1.0cm}<{\centering}p{1.0cm}<{\centering}}
		\hline
		\multirow{2}{*}{Allocation type} & \multicolumn{2}{c}{Old train-set} & \multicolumn{2}{c}{New train-set} \\
		\cline{2-5}
		~& \#images & \#classes & \#images & \#classes \\ 
            \hline 
		Extended-data & 445,419 & 81,313 & 1,580,470 & 81,313 \\
		Extended-class & 470,369 & 24,393 & 1,580,470 & 81,313 \\
		\multirow{2}*{\shortstack{Extended-backbone\\(\textit{class})}} & \multirow{2}{*}{445,419} & \multirow{2}{*}{81,313} & \multirow{2}{*}{1,580,470} & \multirow{2}{*}{81,313}\\\\
		\multirow{2}*{\shortstack{Extended-backbone\\(\textit{data})}} & \multirow{2}{*}{470,369} & \multirow{2}{*}{24,393} & \multirow{2}{*}{1,580,470} & \multirow{2}{*}{81,313}\\\\
		\bottomrule
	\end{tabular}
	\caption{Three different allocations for the training data set sampled from GLDv2. For experiments, the random seed is fixed to reproduce the allocation.}
	\label{tab:allocation}
\end{table}
\section{Experiments}
\subsection{Implementation Details}
\label{sec:impde}
\noindent\textbf{Training Data.} We use Google Landmark v2 \cite{weyand2020google} (GLDv2) as the training dataset. GLDv2 is a large-scale public dataset associated with two challenges Google Landmark Recognition 2019 and Google Landmark Retrieval 2019. Following allocation types mentioned in \cref{sec:benchmark}, the compositions of different training settings are shown in \cref{tab:allocation}.

\noindent\textbf{Training Settings.} All the models are trained on 4 v100 by stochastic gradient descent. For all methods, we train the transformations for 30 epochs, with the learning rate initialized as 0.1. The weight decay is set to 5e-4 and the momentum is 0.9.

\begin{table*}\small
	\centering
	\begin{tabular}{p{0.6cm}<{\centering}p{0.6cm}<{\centering}p{0.6cm}<{\centering}p{0.6cm}<{\centering}|p{1.0cm}<{\centering}p{1.0cm}<{\centering}p{1.0cm}<{\centering}p{1.0cm}<{\centering}|p{1.0cm}<{\centering}p{1.0cm}<{\centering}p{1.0cm}<{\centering}p{1.0cm}<{\centering}}
		\hline
   \multirow{2}{*}{$\#$} & \multirow{2}{*}{$\mathcal{L}_{cls}$} & \multirow{2}{*}{$\mathcal{L}_{adv}$} & \multirow{2}{*}{$\mathcal{L}_{p2s}$} & \multicolumn{2}{c}{RParis} & \multicolumn{2}{c|}{ROxford} & \multicolumn{2}{c}{RParis} & \multicolumn{2}{c}{ROxford} \\ 
			~&~&~&~ & self & cross & self & cross & self & cross& self & cross \\
            \hline 
            1($\phi_o$)& \checkmark & ~&~ &\cellcolor{mygray}75.45 &\cellcolor{mygray} - &\cellcolor{mygray}49.15 &\cellcolor{mygray} - &\cellcolor{mygray}74.29 &\cellcolor{mygray} - &\cellcolor{mygray} 54.34 &\cellcolor{mygray} - \\
			2($\phi_*$)& \checkmark & ~&~ &\cellcolor{mygray} 81.15 &\cellcolor{mygray} 4.93 & \cellcolor{mygray}63.85 & \cellcolor{mygray}1.29 &\cellcolor{mygray}81.15 &\cellcolor{mygray} 4.93 & \cellcolor{mygray}63.85 & \cellcolor{mygray}1.29 \\
			 3&~&  \checkmark& ~& 5.8 &  6.69 &3.25 &1.66&5.14&6.8&2.44&1.76 \\
			 4&~& ~& \checkmark & 76.4 &75.23&50.78&44.75&74.82&74.13&51.09&48.22 \\
			5&\checkmark  &\checkmark & ~&80.87 & 4.4 &\textbf{\textcolor[RGB]{0,0,139}{63.92}} &2.11&81.09&5.67&62.3&2.34 \\
			6&\checkmark  & & \checkmark & \textbf{\textcolor[RGB]{65,105,225}{82.12}}  & \textbf{\textcolor[RGB]{65,105,225}{77.18}}&61.16&\textbf{\textcolor[RGB]{65,105,225}{51.63}}&\textbf{\textcolor[RGB]{65,105,225}{81.66}}&\textbf{\textcolor[RGB]{65,105,225}{76.16}}&\textbf{\textcolor[RGB]{65,105,225}{63.59}}&\textbf{\textcolor[RGB]{65,105,225}{52.92}}\\
			7&~& \checkmark& \checkmark & 76.83 & 75.7&52.76&49.31&75.09&74.45&51.39&48.67\\
			8& \checkmark  &  \checkmark& \checkmark &\textbf{\textcolor[RGB]{0,0,139}{82.78}} & \textbf{\textcolor[RGB]{0,0,139}{78.55}} & \textbf{\textcolor[RGB]{65,105,225}{62.13}} & \textbf{\textcolor[RGB]{0,0,139}{52.31}}&\textbf{\textcolor[RGB]{0,0,139}{82.05}}&\textbf{\textcolor[RGB]{0,0,139}{77.16}}&\textbf{\textcolor[RGB]{0,0,139}{64.51}}&\textbf{\textcolor[RGB]{0,0,139}{54.82}} \\
			\bottomrule
		\end{tabular}
		\caption{Comparison results of different components in Extended-data (left) and Extended-class (right) setting, where both backbones of the old and new model are R18. \textbf{\textcolor[RGB]{0,0,139}{Best}} and \textbf{\textcolor[RGB]{65,105,225}{second best}} are highlighted.}
		\label{tab:each}
    \end{table*}

\noindent\textbf{Evaluation Metrics.} Mean Average Precision ($\textbf{mAP}$) is utilized to evaluate the performance of retrieval. As mentioned in \cref{sec:probfor}, we adopt $\mathcal{P}_{com}$, $\mathcal{P}_{up}$ and $\mathcal{P}_{1-score}$ to evaluate the performance of compatibility and discrimination. We average $\mathcal{P}_{com}$, $\mathcal{P}_{up}$ and $\mathcal{P}_{1-score}$ of every test set as the final $\mathcal{P}_{com}$, $\mathcal{P}_{up}$ and $\mathcal{P}_{1-score}$.

\begin{table*}\small
		\centering
		\begin{tabular}{p{1.4cm}<{\centering}p{1.4cm}<{\centering}p{1.4cm}<{\centering}p{0.7cm}<{\centering}p{0.7cm}<{\centering}p{0.7cm}<{\centering}p{0.7cm}<{\centering}p{0.7cm}<{\centering}p{0.7cm}<{\centering}p{0.8cm}<{\centering}<{\centering}p{0.8cm}<{\centering}p{0.8cm}}
			\toprule			\multirow{2}*{\shortstack{Allocation\\ type}} & Model$_{old}$ & Model$_{new}$ & \multicolumn{2}{c}{RParis} &
			\multicolumn{2}{c}{ROxford} &
			\multicolumn{2}{c}{GLDv2-test} & 
			$\mathcal{P}_{up}$ & $\mathcal{P}_{comp}$ & $\mathcal{P}_{1-score}$ \\
			& & & self & cross & self & cross &
			self & cross & & & \\
			\hline
		    \multirow{5}*{\shortstack{Extended-\\data}} & \cellcolor{mygray}$\phi_o^{R18}$ &\cellcolor{mygray} - &\cellcolor{mygray}75.45  & \cellcolor{mygray}- &\cellcolor{mygray} 49.15 &\cellcolor{mygray} - & \cellcolor{mygray}10.03  &\cellcolor{mygray}-&\cellcolor{mygray}-&\cellcolor{mygray}-&\cellcolor{mygray}- \\
			 & \cellcolor{mygray} - &\cellcolor{mygray} $\phi_{*}^{R18}$ &\cellcolor{mygray} 81.15  &\cellcolor{mygray} 4.93 &\cellcolor{mygray}63.85 &\cellcolor{mygray} 1.20 &\cellcolor{mygray}  16.48 &\cellcolor{mygray} 0.2 &\cellcolor{mygray}-&\cellcolor{mygray}7.19&\cellcolor{mygray}-  \\
			 & $\phi_o^{R18}$ & $\phi_{BCT}^{R18}$ & 80.58&77.37&56.34  & 49.66& 14.61 & 11.30 & 48.02 &54.71&51.13\\
			 & $\phi_o^{R18}$& $\phi_{LCE}^{R18}$ & \textbf{\textcolor[RGB]{65,105,225}{81.57}} & 77.83&\textbf{\textcolor[RGB]{65,105,225}{60.85}}&\textbf{\textcolor[RGB]{65,105,225}{51.35}}&\textbf{\textcolor[RGB]{0,0,139}{16.48}}&\textbf{\textcolor[RGB]{65,105,225}{12.17}}&\textbf{\textcolor[RGB]{0,0,139}{49.65}}&57.41&53.34 \\
			 & $\phi_o^{R18}$& $\phi_{UniBCT}^{R18}$ & 80.93 &\textbf{\textcolor[RGB]{65,105,225}{78.30}} & 57.24&50.97&\textbf{\textcolor[RGB]{65,105,225}{16.06}}&\textbf{\textcolor[RGB]{0,0,139}{13.25}}&48.90&\textbf{\textcolor[RGB]{0,0,139}59.19}&\textbf{\textcolor[RGB]{0,0,139}{53.52}}\\
			 & $\phi_o^{R18}$& $\phi_{Hot-refresh}^{R18}$ & 79.57 &76.53 &58.15&50.05&13.88&10.35&47.78&52.50&50.03 \\
			 & $\phi_o^{R18}$& $\phi_{AdvBCT}^{R18}$ & \textbf{\textcolor[RGB]{0,0,139}{82.78}} & \textbf{\textcolor[RGB]{0,0,139}{78.55}} & \textbf{\textcolor[RGB]{0,0,139}{62.13}} & \textbf{\textcolor[RGB]{0,0,139}{52.31}} & 15.71& 11.49&\textbf{\textcolor[RGB]{65,105,225}{49.55}}&\textbf{\textcolor[RGB]{65,105,225}{58.09}}&\textbf{\textcolor[RGB]{65,105,225}{53.45}} \\
			\hline			\multirow{5}*{\shortstack{Extended-\\class}} & \cellcolor{mygray} $\phi_o^{R18}$ &\cellcolor{mygray} -  & \cellcolor{mygray}74.29  &\cellcolor{mygray} - &\cellcolor{mygray}54.34 &\cellcolor{mygray} - &\cellcolor{mygray}11.43 &\cellcolor{mygray} -&\cellcolor{mygray}-&\cellcolor{mygray}-&\cellcolor{mygray}-\\
			 & \cellcolor{mygray} - &\cellcolor{mygray} $\phi_{*}^{R18}$ &\cellcolor{mygray} 81.15  &\cellcolor{mygray} 4.93 & \cellcolor{mygray}63.85 & \cellcolor{mygray}1.29 & \cellcolor{mygray}16.48 & \cellcolor{mygray}0.2&\cellcolor{mygray} -& \cellcolor{mygray}3.38 &\cellcolor{mygray}-  \\
			 & $\phi_o^{R18}$ & $\phi_{BCT}^{R18}$ & 79.45 & 76.13 & 58.94 & 53.43 &14.79&\textbf{\textcolor[RGB]{65,105,225}{12.26}}&48.33&52.79&50.41 \\
			 & $\phi_o^{R18}$& $\phi_{LCE}^{R18}$ & \textbf{\textcolor[RGB]{65,105,225}{81.26}} & \textbf{\textcolor[RGB]{65,105,225}{76.78}}& \textbf{\textcolor[RGB]{65,105,225}{60.49}}& 54.29&\textbf{\textcolor[RGB]{65,105,225}{16.07}}&12.04&\textbf{\textcolor[RGB]{65,105,225}{49.37}}&53.95&\textbf{\textcolor[RGB]{65,105,225}{51.51}} \\
			 & $\phi_o^{R18}$& $\phi_{UniBCT}^{R18}$ & 76.92 & 74.55 &59.07 &\textbf{\textcolor[RGB]{0,0,139}{57.82}}&14.80&\textbf{\textcolor[RGB]{0,0,139}{12.31}}&48.09&\textbf{\textcolor[RGB]{65,105,225}{54.78}}&51.17 \\
			 & $\phi_o^{R18}$& $\phi_{Hot-refresh}^{R18}$ & 78.93 & 75.33 & 60.31 & 51.68 &14.0&10.41&48.06&47.26&47.57 \\
			 & $\phi_o^{R18}$& $\phi_{AdvBCT}^{R18}$ & \textbf{\textcolor[RGB]{0,0,139}{82.05}} & \textbf{\textcolor[RGB]{0,0,139}{77.16}}&\textbf{\textcolor[RGB]{0,0,139}{64.51}}&\textbf{\textcolor[RGB]{65,105,225}{54.82}}&\textbf{\textcolor[RGB]{0,0,139}{16.44}}&12.05&\textbf{\textcolor[RGB]{0,0,139}{50.16}}&\textbf{\textcolor[RGB]{0,0,139}{54.87}}&\textbf{\textcolor[RGB]{0,0,139}{52.35}} \\
			\hline
			\multirow{5}*{\shortstack{Extended-\\backbone\\(\textit{data})}} &\cellcolor{mygray}  $\phi_o^{R18}$ &\cellcolor{mygray} -  &\cellcolor{mygray} 75.45 & \cellcolor{mygray}- &\cellcolor{mygray} 49.15&\cellcolor{mygray} - &\cellcolor{mygray} 10.03 &\cellcolor{mygray} - &\cellcolor{mygray}-&\cellcolor{mygray}-&\cellcolor{mygray}-\\
			 & \cellcolor{mygray} - &\cellcolor{mygray} $\phi_{*}^{R50}$ &\cellcolor{mygray} 87.66  &\cellcolor{mygray} 4.81 &\cellcolor{mygray} 76.56 & \cellcolor{mygray}2.26&\cellcolor{mygray}  22.12 &\cellcolor{mygray} 0.2&\cellcolor{mygray}-&\cellcolor{mygray}15.44&\cellcolor{mygray}- \\
			 & $\phi_o^{R18}$ & $\phi_{BCT}^{R50}$ &  85.54 & 78.95& 68.66 & 54.42 &19.11 &12.78&47.81&55.86&51.52\\
			 & $\phi_o^{R18}$& $\phi_{LCE}^{R50}$ & \textbf{\textcolor[RGB]{0,0,139}{87.49}} & \textbf{\textcolor[RGB]{0,0,139}{79.36}} & \textbf{\textcolor[RGB]{65,105,225}{75.16}} & 57.18 & \textbf{\textcolor[RGB]{0,0,139}{21.85}} & \textbf{\textcolor[RGB]{65,105,225}{13.30}}&\textbf{\textcolor[RGB]{0,0,139}{49.73}}&\textbf{\textcolor[RGB]{0,0,139}{57.31}}&\textbf{\textcolor[RGB]{0,0,139}{53.25}}\\
			 & $\phi_o^{R18}$& $\phi_{UniBCT}^{R50}$ & 84.37 &78.91 &67.42 & 56.18 & \textbf{\textcolor[RGB]{65,105,225}{21.57}} & \textbf{\textcolor[RGB]{0,0,139}{13.42}}&48.48&56.80&52.31\\
			 & $\phi_o^{R18}$& $\phi_{Hot-refresh}^{R50}$ & 86.78 & 78.81 & 75.10 & \textbf{\textcolor[RGB]{65,105,225}{57.84}} & 20.61 & 12.41&49.19&56.53&52.60 \\
			 & $\phi_o^{R18}$& $\phi_{AdvBCT}^{R50}$ &\textbf{\textcolor[RGB]{65,105,225}{86.79}} & \textbf{\textcolor[RGB]{65,105,225}{79.08}} & \textbf{\textcolor[RGB]{0,0,139}{75.71}}&\textbf{\textcolor[RGB]{0,0,139}{59.03}}&20.77&12.75&\textbf{\textcolor[RGB]{65,105,225}{49.31}}&\textbf{\textcolor[RGB]{65,105,225}{57.30}}&\textbf{\textcolor[RGB]{65,105,225}{53.01}} \\ \hline
			\multirow{5}*{\shortstack{Extended-\\backbone\\(\textit{class})}} & \cellcolor{mygray}$\phi_o^{R18}$ & \cellcolor{mygray}-  & \cellcolor{mygray}74.29 &\cellcolor{mygray} - & \cellcolor{mygray}54.34&\cellcolor{mygray} - & \cellcolor{mygray}11.43 &\cellcolor{mygray} - &\cellcolor{mygray}-&\cellcolor{mygray}-&\cellcolor{mygray}- \\
			 & \cellcolor{mygray}- & \cellcolor{mygray}$\phi_{*}^{R50}$ &\cellcolor{mygray} {87.66}  & \cellcolor{mygray}4.81 &\cellcolor{mygray} {76.56} &\cellcolor{mygray}2.26 &\cellcolor{mygray} 22.12& \cellcolor{mygray}0.2 &\cellcolor{mygray} -&\cellcolor{mygray}11.46&\cellcolor{mygray}-\\
			 & $\phi_o^{R18}$ & $\phi_{BCT}^{R50}$ & 84.18 & 77.12& 68.93 & 57.52 & 18.85&13.47&47.71&54.59&50.91 \\
			 & $\phi_o^{R18}$& $\phi_{LCE}^{R50}$ & \textbf{\textcolor[RGB]{65,105,225}{85.77}} & 77.31 & 72.89 & \textbf{\textcolor[RGB]{65,105,225}{58.95}}&\textbf{\textcolor[RGB]{65,105,225}{20.84}}&\textbf{\textcolor[RGB]{0,0,139}{14.00}}&\textbf{\textcolor[RGB]{65,105,225}{49.04}}&\textbf{\textcolor[RGB]{65,105,225}{55.67}}&\textbf{\textcolor[RGB]{65,105,225}{52.15}} \\
			 & $\phi_o^{R18}$& $\phi_{UniBCT}^{R50}$ & 82.48 & \textbf{\textcolor[RGB]{65,105,225}{77.33}}&66.37&58.51&18.91&\textbf{\textcolor[RGB]{65,105,225}{13.99}}&47.29&55.51&51.06 \\
			 & $\phi_o^{R18}$& $\phi_{Hot-refresh}^{R50}$ & 85.62 &77.29& \textbf{\textcolor[RGB]{0,0,139}{73.95}} &56.83&20.47&12.44&49.01&53.63&51.21\\
			 & $\phi_o^{R18}$& $\phi_{AdvBCT}^{R50}$ & \textbf{\textcolor[RGB]{0,0,139}{86.24}} & \textbf{\textcolor[RGB]{0,0,139}{78.31}} & \textbf{\textcolor[RGB]{65,105,225}{73.93}} & \textbf{\textcolor[RGB]{0,0,139}{60.33}} & \textbf{\textcolor[RGB]{0,0,139}{22.03}} & 13.54&\textbf{\textcolor[RGB]{0,0,139}{49.65}}&\textbf{\textcolor[RGB]{0,0,139}{56.64}}&\textbf{\textcolor[RGB]{0,0,139}{52.83}} \\
			\bottomrule
		\end{tabular}
		\caption{The compatible training benchmark testing on \textit{BCT}, \textit{LCE}, \textit{Hot-refresh}, \textit{UniBCT}, and \textit{AdvBCT}. $\phi_o^{R18}$ represents that the backbone of the old model is ResNet18. $\phi_{*}$ is trained on the whole data set without any compatible operation. $self$ represents self-test $M(\phi_{o},\phi_{o};\mathcal{Q},\mathcal{G})$ or $M(\phi_{n},\phi_{n};\mathcal{Q},\mathcal{G})$, and $cross$ represents cross-test $M(\phi_{n},\phi_{o};\mathcal{Q},\mathcal{G})$. In the allocation type of Extended-backbone, \textit{data} represents 30\% and 100\% training data for the old model and the new model respectively. Similarly, \textit{class} represents 30\% class and 100\% class for the old model and the new model. Our method \textit{AdvBCT} achieves best performance in the most allocation types. \textbf{\textcolor[RGB]{0,0,139}{Best}} and \textbf{\textcolor[RGB]{65,105,225}{second best}} of five works are higlighted.}
		\label{tab:benchmark}
	\end{table*}



\subsection{Ablation Study}
	\noindent\textbf{Effectiveness of different components.} As shown in \cref{fig:method}, our \textit{AdvBCT} has three modules. To verify the impact of each module alone on the final effect, we conduct eight split experiments on each module under the compatible settings of Extended-data and Extended-class, and the effect is evaluated on RParis and ROxford datasets. The experimental results are shown in \cref{tab:each}. 
 
 As can be seen from the table, each component is necessary, and all components are combined to achieve the best overall effect. In the first four sets of experiments, each component is tested individually. As shown in \cref{tab:each}, $\mathcal{L}_{cls}$ makes the new model discriminative but lacking compatibility (refer to $\#1$ and $\#2$), and $\mathcal{L}_{p2s}$ performs better compatibility but limited discrimination (refer to $\#4$, $\#2$ and $\#1$). In addition, we also found that the effect of $\mathcal{L}_{adv}$ alone is relatively poor. It's reasonable 
that adversarial learning is an unsupervised constraint on old and new distributions which cannot constrain instances directly. But when $\mathcal{L}_{adv}$ and $\mathcal{L}_{p2s}$ are combined, better results are achieved (refer to $\#4$ and $\#7$). Of course, from the experimental results, the combination of the three achieves the best discrimination and compatibility. Therefore, our proposed adversarial and boundary-aware compatible modules are all effective.

\begin{figure}
        \centering
        \includegraphics[width=0.8\linewidth]{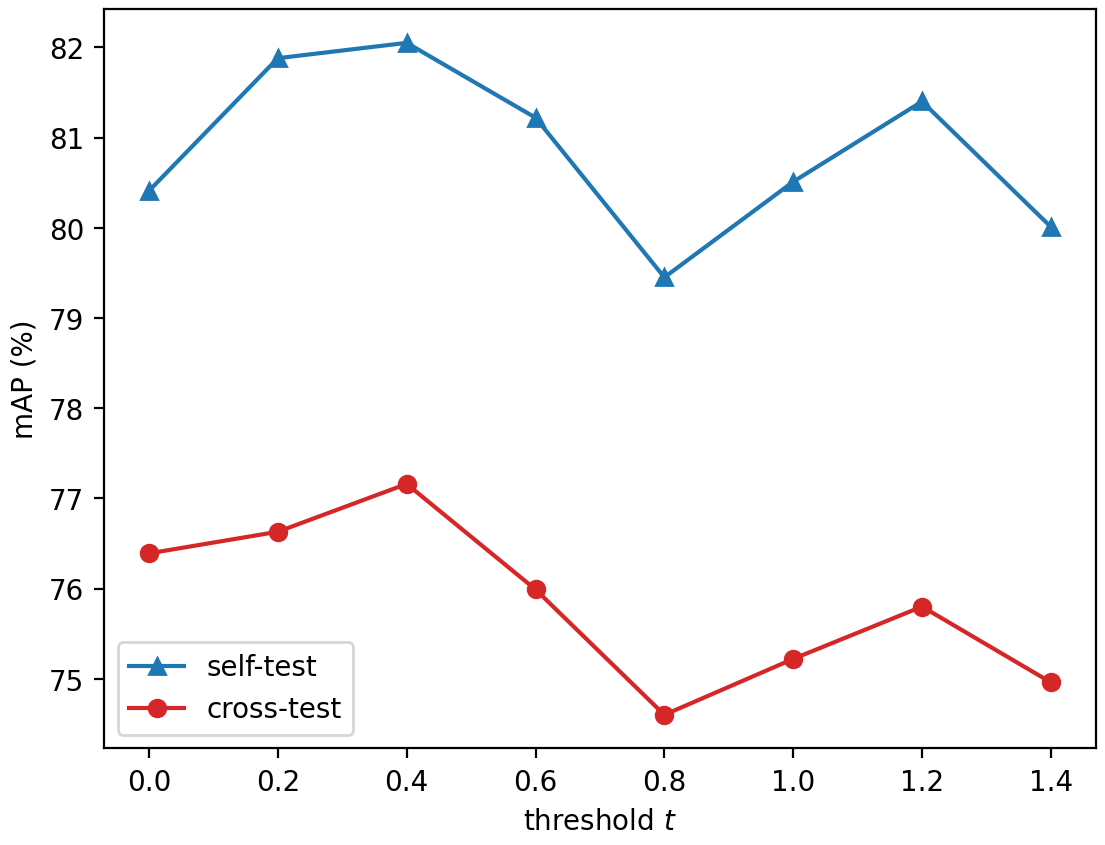}
        \caption{The influence of parameter $t$ on RParis dataset in Extended-class setting.}
        \label{fig:param}
\end{figure}

\noindent\textbf{Parameter Analysis.} In order to determine an appropriate boundary to constrain the distance between the new embeddings and the old cluster centers, we set a flexible threshold $t$ to obtain the upper or the lower of boundaries in formula \cref{equ:rmax}. To observe the influence of threshold $t$, we conduct experiments on RParis dataset in the Extended-class setting. The retrieval performance of different $t$ is shown in figure \cref{fig:param}. When $t=0.4$, we get the best mAP on both self-test and cross-test. From the figure, we can see that large thresholds and small thresholds perform worse. When the threshold is greater, the model has more flexibility to learn discriminative embeddings, but the compatibility is limited. On the contrary, when the constraint on compatibility is too strict, discrimination is affected. 
	
\begin{figure}
        \centering
        \includegraphics[width=0.8\linewidth]{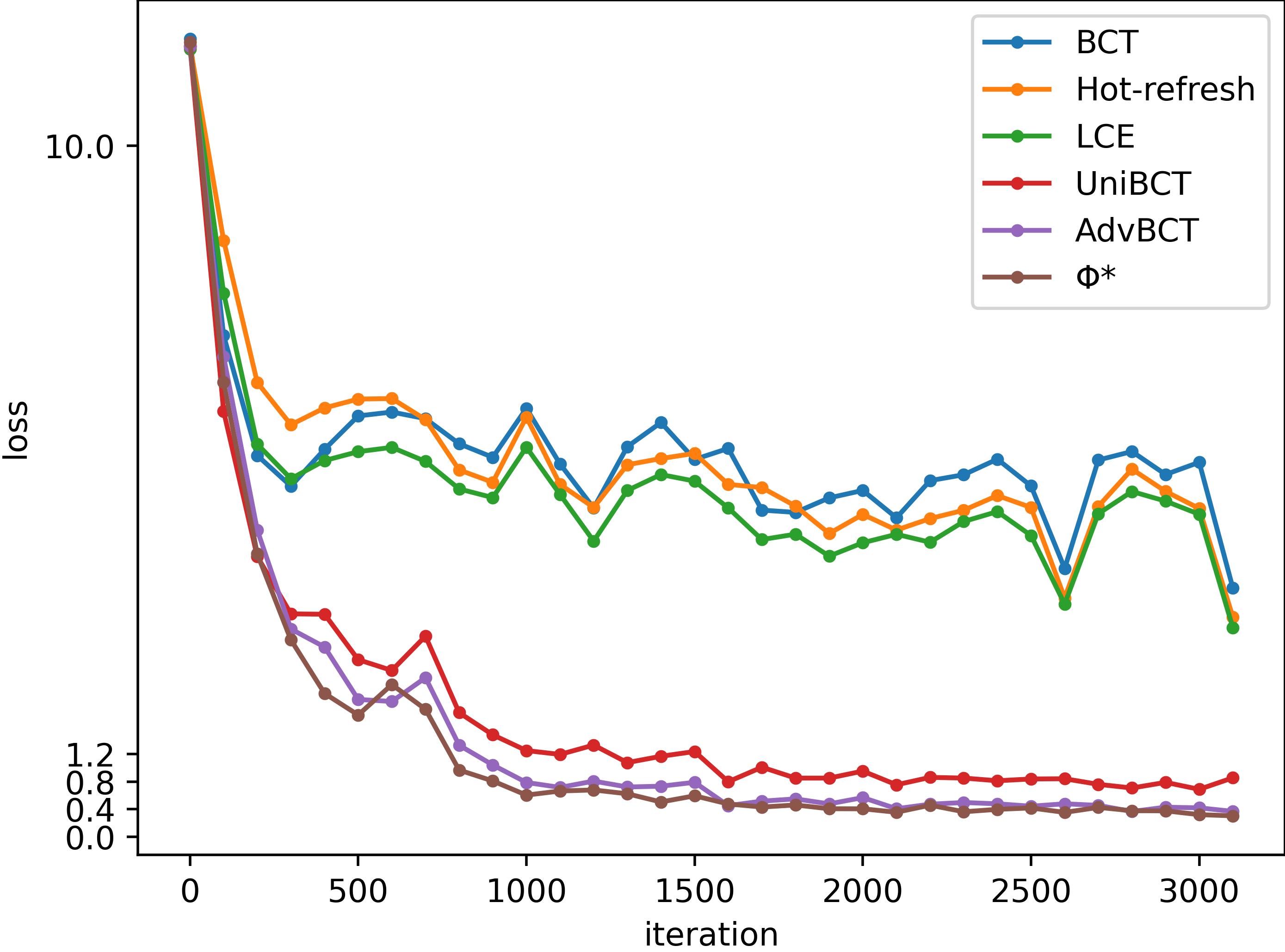}
        \caption{The convergence trend of $\textit{BCT}$\cite{shen2020towards}, $\textit{LCE}$\cite{meng2021learning}, $\textit{Hot-refresh}$\cite{zhang2021hot}, $\textit{UniBCT}$\cite{zhang2022towards}, our \textit{AdvBCT} and $\phi_{*}$ in Extended-backbone (class) setting.}
        \label{fig:lossvisual}
\end{figure}

\subsection{Compatible Training Benchmark}
Following benchmark settings mentioned in \cref{sec:benchmark}, we evaluate 4 previous works labeled as $\textit{BCT}$\cite{shen2020towards}, $\textit{LCE}$\cite{meng2021learning}, $\textit{Hot-refresh}$\cite{zhang2021hot}, and $\textit{UniBCT}$\cite{zhang2022towards} and our work $\textit{AdvBCT}$. Thanks to the open source of the Ref\cite{zhang2021hot}, we implemented $\textit{BCT}$ and  $\textit{Hot-refresh}$ according to their work. We implemented $\textit{UniBCT}$ without the structural prototype refinement algorithm and followed the setting in $\textit{LCE}$ that the transformation layer $K=0$ while compatible learning. We mark $M(\phi_{n},\phi_{n};\mathcal{Q},\mathcal{G})$ or $M(\phi_{o},\phi_{o};\mathcal{Q},\mathcal{G})$ as \textit{self-test} and $M(\phi_{n},\phi_{o};\mathcal{Q},\mathcal{G})$ as \textit{cross-test}.

The results of different allocation types are shown in \cref{tab:benchmark}. From the experiment results, our $\textit{AdvBCT}$ surpasses previous BCT works in the most cases. Especially in the scenario that classes increase, $\textit{AdvBCT}$ exceeds other methods more than 1\% in $\mathcal{P}_{1-score}$.

It is remarkable that the self-test performance of $\textit{AdvBCT}$ and $\textit{LCE}$ outperforms the performance of $\phi_*$ in Extended-data and Extended-class. The results indicate that the compatibility has positive impacts on discrimination of classes, if the constraints are selected properly. Furthermore, from the results, the method $\textit{UniBCT}$ which minimizes the distances of the old prototypes and new embeddings directly performs not as good as the methods $\textit{LCE}$ and $\textit{AdvBCT}$ in self-test whose constraints are under boundary limits. That means boundary limits are meaningful for improvements on the self-test.

We also visualize convergence trend of several methods in \cref{fig:lossvisual}. From the figure, we can see that our method converges fast, the loss declined smooth and the trend is close to $\phi_*$. One explanation is that adversarial learning pulls two models into the same distribution which can be helpful to the distance compatible constraint. And the elastic boundary gives more freedom space to optimizing compatibility and discrimination.

\section{Conclusion}
In this paper, we proposed a novel backward-compatible training method in image retrieval. To better ensure compatibility, we designed the adversarial and boundary-aware compatible modules. Adversarial compatible module aims to pull the embedding distributions of the old and new models close. And boundary-aware compatible module is used to obtain a suitable boundary to constrain distance relationship between the new and old embeddings. 
In addition, we compare our \textit{AdvBCT} with the existing BCT methods in uniform settings, and an eclectic metric is proposed to verify the pros and cons of all backward-compatible methods, which establishes a comprehensive benchmark for subsequent researchers to handily contribute to the field. Extensive experiments were conducted to verify the effectiveness of our \textit{AdvBCT}. For our future work, we will explore leveraging the old embeddings to further improve discrimination while maintaining compatibility.
\section*{Acknowledgments}
This work is partly supported by R\&D Program of DCI Technology and Application Joint Laboratory.
{\small
\bibliographystyle{ieee_fullname}
\bibliography{egbib}

\begin{thebibliography}{10}\itemsep=-1pt

\bibitem{duggal2021compatibility}
Rahul Duggal, Hao Zhou, Shuo Yang, Yuanjun Xiong, Wei Xia, Zhuowen Tu, and
  Stefano Soatto.
\newblock Compatibility-aware heterogeneous visual search.
\newblock In {\em CVPR}, pages 10723--10732, 2021.

\bibitem{ganin2015unsupervised}
Yaroslav Ganin and Victor Lempitsky.
\newblock Unsupervised domain adaptation by backpropagation.
\newblock In {\em ICML}, pages 1180--1189, 2015.

\bibitem{ge2020self}
Yixiao Ge, Feng Zhu, Dapeng Chen, Rui Zhao, et~al.
\newblock Self-paced contrastive learning with hybrid memory for domain
  adaptive object re-id.
\newblock In {\em NIPS}, pages 11309--11321, 2020.

\bibitem{goodfellow2020generative}
Ian Goodfellow, Jean Pouget-Abadie, Mehdi Mirza, Bing Xu, David Warde-Farley,
  Sherjil Ozair, Aaron Courville, and Yoshua Bengio.
\newblock Generative adversarial networks.
\newblock {\em CACM}, 63(11):139--144, 2020.

\bibitem{hu2018duplex}
Lanqing Hu, Meina Kan, Shiguang Shan, and Xilin Chen.
\newblock Duplex generative adversarial network for unsupervised domain
  adaptation.
\newblock In {\em CVPR}, pages 1498--1507, 2018.

\bibitem{jiang2021transgan}
Yifan Jiang, Shiyu Chang, and Zhangyang Wang.
\newblock Transgan: Two pure transformers can make one strong gan, and that can
  scale up.
\newblock In {\em NIPS}, pages 14745--14758, 2021.

\bibitem{karras2019style}
Tero Karras, Samuli Laine, and Timo Aila.
\newblock A style-based generator architecture for generative adversarial
  networks.
\newblock In {\em CVPR}, pages 4401--4410, 2019.

\bibitem{li2011towards}
Beibei Li, Anindya Ghose, and Panagiotis~G Ipeirotis.
\newblock Towards a theory model for product search.
\newblock In {\em WWW}, pages 327--336, 2011.

\bibitem{lin2019real}
Kai Lin, Thomas~H Li, Shan Liu, and Ge Li.
\newblock Real photographs denoising with noise domain adaptation and attentive
  generative adversarial network.
\newblock In {\em CVPR}, pages 0--0, 2019.

\bibitem{long2018conditional}
Mingsheng Long, Zhangjie Cao, Jianmin Wang, and Michael~I Jordan.
\newblock Conditional adversarial domain adaptation.
\newblock volume~31, 2018.

\bibitem{matthews1975comparison}
Brian~W Matthews.
\newblock Comparison of the predicted and observed secondary structure of t4
  phage lysozyme.
\newblock {\em BBA-PS}, 405(2):442--451, 1975.

\bibitem{maze2018iarpa}
Brianna Maze, Jocelyn Adams, James~A Duncan, Nathan Kalka, Tim Miller, Charles
  Otto, Anil~K Jain, W~Tyler Niggel, Janet Anderson, Jordan Cheney, et~al.
\newblock Iarpa janus benchmark-c: Face dataset and protocol.
\newblock In {\em ICB}, pages 158--165, 2018.

\bibitem{mei20203rd}
Ke Mei, Jinchang Xu, Yanhua Cheng, Yugeng Lin, et~al.
\newblock 3rd place solution to" google landmark retrieval 2020".
\newblock {\em arXiv preprint arXiv:2008.10480}, 2020.

\bibitem{meng2021learning}
Qiang Meng, Chixiang Zhang, Xiaoqiang Xu, and Feng Zhou.
\newblock Learning compatible embeddings.
\newblock In {\em CVPR}, pages 9939--9948, 2021.

\bibitem{ramanujan2022forward}
Vivek Ramanujan, Pavan Kumar~Anasosalu Vasu, Ali Farhadi, Oncel Tuzel, and Hadi
  Pouransari.
\newblock Forward compatible training for large-scale embedding retrieval
  systems.
\newblock In {\em CVPR}, pages 19386--19395, 2022.

\bibitem{rowley2000product}
Jennifer Rowley.
\newblock Product search in e-shopping: a review and research propositions.
\newblock {\em JCM}, 2000.

\bibitem{schroff2015facenet}
Florian Schroff, Dmitry Kalenichenko, and James Philbin.
\newblock Facenet: A unified embedding for face recognition and clustering.
\newblock In {\em CVPR}, pages 815--823, 2015.

\bibitem{shen2020towards}
Yantao Shen, Yuanjun Xiong, Wei Xia, and Stefano Soatto.
\newblock Towards backward-compatible representation learning.
\newblock In {\em CVPR}, pages 6368--6377, 2020.

\bibitem{srivastava2020empirical}
Megha Srivastava, Besmira Nushi, Ece Kamar, Shital Shah, and Eric Horvitz.
\newblock An empirical analysis of backward compatibility in machine learning
  systems.
\newblock In {\em SIGKDD}, pages 3272--3280, 2020.

\bibitem{van2008visualizing}
Laurens Van~der Maaten and Geoffrey Hinton.
\newblock Visualizing data using t-sne.
\newblock {\em JMLR}, 9(11), 2008.

\bibitem{wang2018deep}
Mei Wang and Weihong Deng.
\newblock Deep visual domain adaptation: A survey.
\newblock {\em Neurocomputing}, 312:135--153, 2018.

\bibitem{weyand2020google}
Tobias Weyand, Andre Araujo, Bingyi Cao, and Jack Sim.
\newblock Google landmarks dataset v2-a large-scale benchmark for
  instance-level recognition and retrieval.
\newblock In {\em CVPR}, pages 2575--2584, 2020.

\bibitem{wu2022neighborhood}
Shengsen Wu, Liang Chen, Yihang Lou, Yan Bai, Tao Bai, Minghua Deng, and
  Ling-Yu Duan.
\newblock Neighborhood consensus contrastive learning for backward-compatible
  representation.
\newblock In {\em AAAI}, volume~36, pages 2722--2730, 2022.

\bibitem{xu20213rd}
Cheng Xu, Weimin Wang, Shuai Liu, Yong Wang, Yuxiang Tang, Tianling Bian, Yanyu
  Yan, Qi She, and Cheng Yang.
\newblock 3rd place solution to google landmark recognition competition 2021.
\newblock {\em arXiv preprint arXiv:2110.02794}, 2021.

\bibitem{xu2021prdp}
Furong Xu, Bingpeng Ma, Hong Chang, and Shiguang Shan.
\newblock Prdp: Person reidentification with dirty and poor data.
\newblock {\em ToC}, 2021.

\bibitem{xu2018style}
Furong Xu, Bingpeng Ma, Hong Chang, Shiguang Shan, and Xilin Chen.
\newblock Style transfer with adversarial learning for cross-dataset person
  re-identification.
\newblock In {\em ACCV}, pages 165--180, 2018.

\bibitem{xu2021discrimination}
Furong Xu, Meng Wang, Wei Zhang, Yuan Cheng, and Wei Chu.
\newblock Discrimination-aware mechanism for fine-grained representation
  learning.
\newblock In {\em CVPR}, pages 813--822, 2021.

\bibitem{xu2020metric}
Furong Xu, Wei Zhang, Yuan Cheng, and Wei Chu.
\newblock Metric learning with equidistant and equidistributed triplet-based
  loss for product image search.
\newblock In {\em WWW}, pages 57--65, 2020.

\bibitem{you2019universal}
Kaichao You, Mingsheng Long, Zhangjie Cao, Jianmin Wang, and Michael~I Jordan.
\newblock Universal domain adaptation.
\newblock In {\em CVPR}, pages 2720--2729, 2019.

\bibitem{zhang2021hot}
Binjie Zhang, Yixiao Ge, Yantao Shen, Yu Li, Chun Yuan, Xuyuan Xu, Yexin Wang,
  and Ying Shan.
\newblock Hot-refresh model upgrades with regression-free compatible training
  in image retrieval.
\newblock In {\em ICLR}, 2021.

\bibitem{zhang2022towards}
Binjie Zhang, Yixiao Ge, Yantao Shen, Shupeng Su, Chun Yuan, Xuyuan Xu, Yexin
  Wang, and Ying Shan.
\newblock Towards universal backward-compatible representation learning.
\newblock 2020.

\bibitem{zhao2003face}
Wenyi Zhao, Rama Chellappa, P~Jonathon Phillips, and Azriel Rosenfeld.
\newblock Face recognition: A literature survey.
\newblock {\em CSUR}, 35(4):399--458, 2003.

\bibitem{zheng2019joint}
Zhedong Zheng, Xiaodong Yang, Zhiding Yu, Liang Zheng, Yi Yang, and Jan Kautz.
\newblock Joint discriminative and generative learning for person
  re-identification.
\newblock In {\em CVPR}, pages 2138--2147, 2019.

\bibitem{zhou2022forward}
Da-Wei Zhou, Fu-Yun Wang, Han-Jia Ye, Liang Ma, Shiliang Pu, and De-Chuan Zhan.
\newblock Forward compatible few-shot class-incremental learning.
\newblock In {\em CVPR}, pages 9046--9056, 2022.

\end{thebibliography}
}

\end{document}